\documentclass{ecai}
\usepackage{xspace}
\usepackage{graphicx}
\usepackage{graphics}
\usepackage[numbers]{natbib}
\usepackage{hyperref}
\usepackage{graphicx,amsmath,amssymb}
\usepackage[ruled,linesnumbered]{algorithm2e}
\usepackage{enumerate}
\usepackage{slantsc}
\makeatletter
\newcommand\footnoteref[1]{\protected@xdef\@thefnmark{\ref{#1}}\@footnotemark}
\makeatother
\ecaisubmission   
\newboolean{showcomments}
\setboolean{showcomments}{false}
\ifthenelse{\boolean{showcomments}}
  {\newcommand{\nb}[3]
    {
    {\color{#2}\small\fbox{\bfseries\sffamily\scriptsize#1}}
    {\color{#2}\sffamily\small$\triangleright~$\textit{\small #3}$~\triangleleft$}
    }
  }
  {    \newcommand{\nb}[3]{}  }
\newcommand\rmr[1]{\nb{\textbf{Reshef:}}{blue}{#1}}

\def\BIS{BIS\xspace}
\def\IO{IO\xspace}
\def\MS{MS\xspace}
\def\RB{RB\xspace}
\def\SRP{SRP\xspace}
\def\WRP{WRP\xspace}
\def\UCT{UCT\xspace}
\newcommand{\ol}[1]{\overline{#1}}
\newcommand{\oll}[1]{\ol{\ol{#1}}}
\renewcommand{\paragraph}[1]{\medskip\noindent{\bfseries {#1.}}}
\begin{document}
\title{Bidding in Spades}
\author{Gal Cohensius\institute{Technion, Israel, email: galcohensius@campus.technion.ac.il} \and Reshef Meir\institute{Technion, Israel, email: reshefm@ie.technion.ac.il} \and Nadav Oved\institute{Technion, Israel, email: nadavo@campus.technion.ac.il} \and Roni Stern\institute{Ben Gurion University, Israel, email: sternron@post.bgu.ac.il, Palo Alto Research Center (PARC), CA, USA} 
}
\maketitle
 \bibliographystyle{ecai}
\begin{abstract}
    We present a Spades bidding algorithm that is superior to recreational human players and to publicly available bots.  Like in Bridge, the game of Spades is composed of two independent phases, \textit{bidding} and \textit{playing}.  This paper focuses on the bidding algorithm, since this phase holds a precise challenge: based on the input, choose the bid that maximizes the agent's winning probability.  
    Our \emph{Bidding-in-Spades} (\BIS) algorithm heuristically determines the bidding strategy by comparing the expected utility of each possible bid. 
    A major challenge is how to estimate these expected utilities. 
    To this end, we propose a set of domain-specific heuristics, and
    then correct them via machine learning using data from real-world players. 
    The \BIS algorithm we present can be attached to any playing algorithm.  It beats rule-based bidding bots when all use the same playing component. When combined with a rule-based playing algorithm,  it is superior to the average recreational human.
\end{abstract}

\section{Introduction}
Spades is a popular card game. Therefore designing strong Spades agents has a commercial value, as millions of games are held daily on mobile applications. Those applications have been downloaded from Google Play store more than any other trick taking game (over 10M times) and produce annual income of several millions of dollars\cite{10-Best-Card-Games-for-Android-in-2020, best-android-card-games}.

Spades shares many similarities with games such as Bridge, Skat, Hearts and others that have attracted considerable attention in AI.  Recently, AI agents have reached superhuman performance in the partial information game of no-limit poker~\cite{brown2018superhuman,brown2019superhuman} and to a level of human experts in Bridge, which is considered to be one of the most appraised partial information problems for AI~\cite{amit2006learning, yeh2018automatic, rong2019competitive}.  Spades, however, has received relatively little attention in the literature.  Although several Spades bots were made, such as those made by AI-factory (see Related Work),  
we have no knowledge of strong \emph{publicly available} bots, thus a comparison with the state of the art algorithm is unavailable.  Instead, we compared our bidding in Spades (\BIS) to humans and to rule-based bidding bots on one of the most popular mobile applications and show that \BIS bidding is superior. 

The game holds three interesting features from AI perspective: (1) It is a two-versus-two game, meaning that each agent has a partner and two opponents.  The partner can be an AI ``friend" with a common signal convention or an unknown AI/human where no convention can be assumed; 
(2) Partly observable state: agents observe their hand but do not know how the remaining cards are distributed between the other players.  Each partly observable state at the start of a round can be completed to a full state in $\frac{39!}{13!^3}\cong 8.45\cdot10^{16}$ ways; and
(3) Goal choosing, as different bids mean that the agent should pursue different goals during the round.  

\paragraph{Related work} We first mention two general game-playing algorithms: \emph{Monte-Carlo Tree Search} (MCTS) evaluates moves by simulating many random games and taking the average score~\cite{brugmann1993monte}. \emph{Upper Confidence  bounds applied to Trees} (\UCT) is an improved version that biases the sampling in favor of moves that already have higher score~\cite{kocsis2006bandit}.

Two groups have made intensive research in the specific area of Spades agents: a group from University of Alberta~\cite{sturtevant2002comparison, sturtevant2008analysis, sturtevant2006robust, sturtevant2006prob} and AI-Factory group \cite{baier2018emulating, cowling2012information,  devlin2016combining, whitehouse2013integrating}.  The latter launched  a commercial application called Spades Free.  AI-Factory use a knowledge-based bidding module because they found that a ``Monte Carlo Tree Search is poor at making bidding decisions''~\cite{whitehouse2013integrating}.  
Whitehouse et al.~\cite{whitehouse2013integrating} presented an improved MCTS playing module that beats their strongest heuristic playing agent, they explain that the MCTS must use several heuristics tweaks to be perceived as strong by human players.  
In a follow-up study, Baier et al.~\cite{baier2018emulating} used neural networks in order to emulate human play. They 
trained the network to predict human's next moves, given a game state. It achieved an accuracy of $67.8\%$ which is an improvement over other techniques such as decision trees.

The Alberta University team considered a simplified version of the game with perfect-information (hands are visible), 3 players, 7 cards, and no partnerships.  This reduction was essential in order to reduce the size of possible states of the game, which allow search algorithms to get good results faster.
Sturtevant et al.~\cite{sturtevant2002comparison} compared the \textit{paranoid}~\cite{sturtevant2000pruning} and the \textit{$\max^n$}~\cite{luckhart1986algorithmic} algorithms to a handmade heuristic and found that both algorithms were barely able to outperform their handmade heuristic. 
A  followup research from the same group showed that in an n-player game, search algorithms must use \textit{opponents modeling} in order to obtain good results~\cite{sturtevant2006robust}.  They proposed the \textit{soft-$\max^n$} algorithm which uses opponent modeling.  Interestingly, they used a rule based bidding system, even though the search space is much smaller than classic Spades. 
In the simple 3-player Spades variant mentioned above,  \UCT reaches the level of play of  prob-$\max^n$~\cite{sturtevant2008analysis}.  Authors hypothesized that \UCT works better in games with high branching factor and low n-ply variance\footnote{A measure of how fast the game state can change.}  such as Gin-Rummy.  

Besides those two groups, dozens of Spades apps are available on Google's play store, most of them have an option of playing with AI players.  Currently, The most popular apps are Zinga's Spades Plus,\footnote{www.zynga.com/games/spades-plus} BeachBum's Spades Royale\footnote{www.spadesroyale.com} and AI Factory's Spades Free.\footnote{www.aifactory.co.uk}

\paragraph{Contribution}
The main focus of the \BIS agent is the decision whether to bid \emph{nil}. \BIS uses Monte Carlo simulations combined with heuristic relaxations to obtain heuristic values for nil and non-nil bids. The  probability of winning a nil bid is then evaluated using supervised learning from millions of real games. 

Combined with a rule-based playing module, \BIS is superior to other rule-based bidding algorithms and to the average recreational human, beating humans in 56\% of the games. In particular,  \BIS bids nil more frequently ($13.6\%$ of the rounds vs. $12.6\%$), and still obtains a higher success rate in those rounds ($68.8\%$ vs.  $63.2\%$ for human players).

\begin{figure}[h]
    \centerline{ \includegraphics[width=\linewidth, height = 100pt]{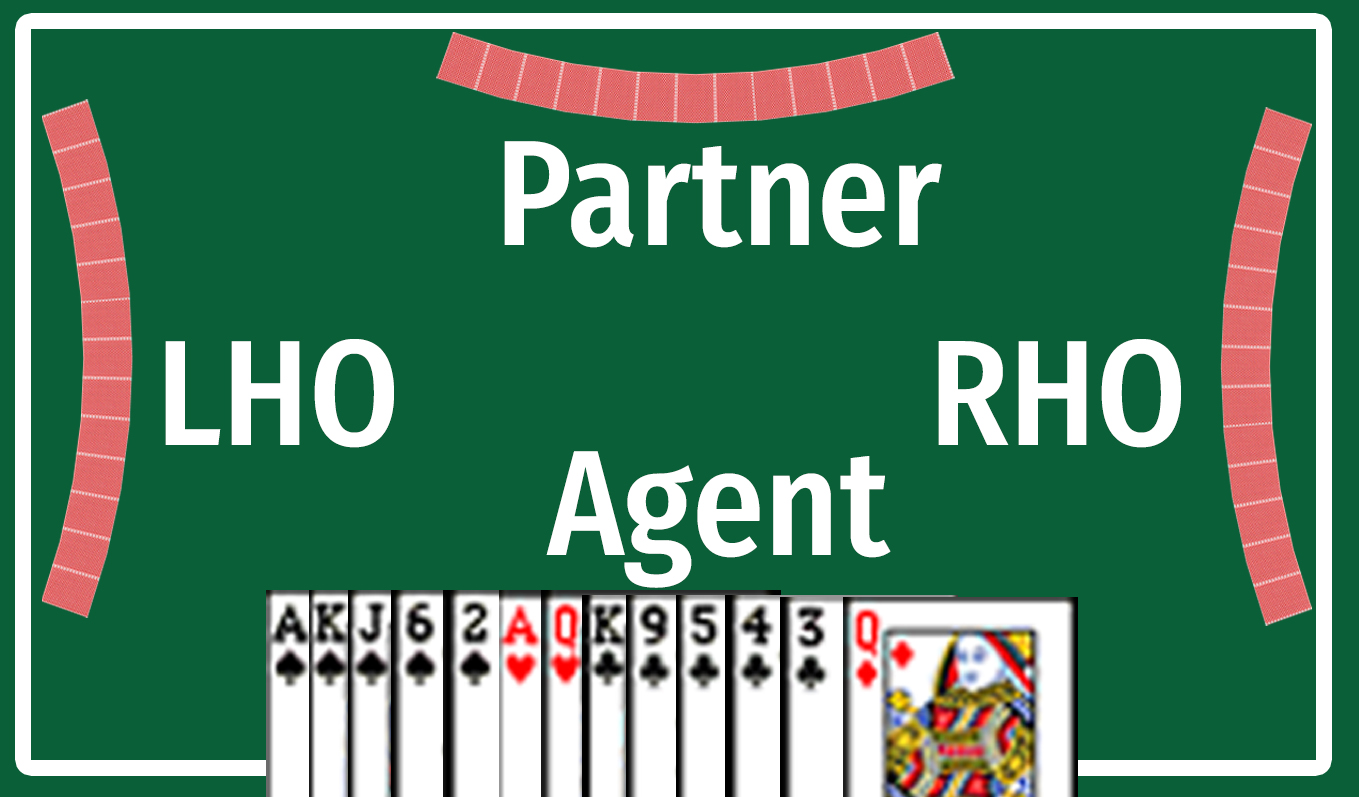} }
    \caption{A game of Spades. From the agent's perspective, the Left-Hand Opponent (LHO), Right-Hand Opponent (RHO) and the Partner are sitting at West, East and North, respectively.} \label{fig:table}
\end{figure}
\vspace{-6mm}
\section{Rules of Spades}
Spades is a 4-player trick-taking card game.  It resembles Whist, Euchre, Oh Hell, Hearts, Skat, Callbreak and often referred to as a simpler version of Bridge.  Its name comes from the rule that the spade suit is strongest (spades are trumps).  The game is played in partnerships of two, partners sit across the table from each other and named after the four compass directions: East and West against North and South.  The game is played over several \textit{rounds}.  At the end of each round both partnerships score points, the winner is the partnership with the highest score that also exceeds a predefined winning goal (usually 500 points).   A round begins with dealing 13 cards to each player out of a regular deck of 52 cards.   Each round has two phases: \textit{bidding} followed by \textit{playing}.

In the bidding phase, each player, in her turn (passed clockwise), makes a single \textit{bid} (0-13), which states the number of \textit{tricks} she commits to take. The playing phase has 13 tricks, where a trick consists of each player in her turn, playing a single card from her hand onto the table.  The player which played the strongest card on the table, wins the trick and will be the leader of the next trick.  The first card played in a trick is the leading card and determines the \textit{leading suit} of the trick, other players must follow the leading suit if they can.  If in a trick, no spade cards were played, then the highest \textit{leading suit} card is the winner, if a spade card was played then the highest spade card is the winner.  The leader cannot lead a spade card unless \textit{spades were broken} or she is \textit{spades tight} (holds nothing but spades).  Spades are \textit{broken} when a spade card is played for the first time in the round.  

When a round ends, scoring is performed.  A partnership that takes at least as many tricks as the combined bid of both partners, receives 10 points times their combined bid, otherwise the partnership loses 10 points times their combined bid.  Any extra trick taken beyond their combined bid is called \textit{bag} (or overtrick) and it is worth 1 point.  If throughout the game a partnership collects 10 bags, they lose 100 points and 10 bags.  Thus players usually aim to take exactly their bid.  

For example, assume agent bids $4$ and partner bids $2$.  If their sum of takes is less than $6$ tricks then they will lose $60$ points, if they will take $9$ tricks, then they will receive $63$ points.  If they had already $288$ points (meaning they collected 8 bags during previous rounds), then they will receive $63$ however since they cross the $10$ bags limit they will lose $10$ bags and $100$ points, which will results in $241$ points.

Bidding 0, also known as  \textit{nil}, is a special bid.  If a player bids nil then each of the partners in the partnership checks separately whether her bid was successful.  A player that bids nil wins 100 points if she individually did not win any trick, and loses 100 points if she did.    
Terminology can be found in appendix\ref{app: Terminology}.  The complete set of rules can be found at The Pagat website~\cite{Paget}.

\section{To Nil or Not to Nil, that is The Question}
\begin{figure}
    \centerline{ \includegraphics[width=0.8\linewidth]{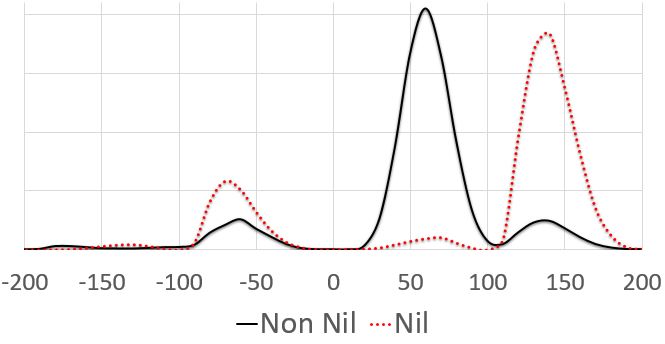} }
    \caption{Histograms of the points gained in a round by \BIS{}+Partner, when agent bids either nil or non-nil. The main peak of the ``non-nil" histogram (around $60$ points) is due to agents fulfilling their regular bid. The left peak ($-60$) is due to failed bids, and the right peak ($130$) is due to successful nil bids of the partner.} \label{fig:points_per_round_GG}
\end{figure}

The major decision a player is facing during the bidding phase is whether to bid nil or a regular bid (non nil bid).  A nil bid offers a high reward ($100$ points) but a high risk of being set ($-100$ points) compared to a regular bid, making the decision a risk-reward trade-off.
The major factor of nil bids is shown at Fig.~\ref{fig:points_per_round_GG}. While the score of no-nil rounds is concentrated around 60 points, nil bids result in a risky gamble (see the two peaks of the ``Nil" curve). 

Deciding to nil does not depend solely on the hand a player holds. 
The relevant parameters are: (1) the hand's compatibility to a nil bid; (2) the alternative regular bid; (3) the bids by previous players; and (4) the willingness to take risks, which depends on the current scores in the game.  For example when the other partnership is about to win the game while having an easy contract, a desperate nil might be the best option.  High alternative regular bid reduces the risk a player should take since the alternative reward is higher.  Bidding later in the turn order reveals information about the other players hands. The most important information is the partner's bid.  High partner's bid informs that she holds several high cards, which allows the agents to bid nil even with hand containing high cards.  A high sum of bids by the three other players reduces the risk for nil since the opponents cannot set the nil without jeopardizing their own bid.

\section{The \BIS Agent}
The \BIS agent runs two preprocesses: the first is a Monte-Carlo simulation that estimates the probability that other players hold various cards. The output of these simulations is in the form of Probability Tables, that we denote by PT. 
The second preprocess uses data from real games played against human players, to estimate success probabilities of nil bids. The output is in the form a real-valued function called \emph{success curves} (SC).

In every round, the bid is determined in the following manner. The agent may use the precomupted data (PT and SC), in addition to the current hand and the sequence of 0-3 previous bids in this round.  

\begin{algorithm}[ht]
    
    \SetAlgoLined
    Input: Hand, PrevBids\;
    \KwResult{ bid [0-13]}
    regularTakes $\leftarrow$ CalcRegularTakes(PT,Hand,PrevBids)\;
    nilValue $\leftarrow$ CalcNilValue(PT,Hand)\;
    nilProb $\leftarrow$ SC(PrevBids,nilValue)\;
    expNilScore $\leftarrow$ nilProb $\cdot 100$ + (1-nilProb) (-100)\;
    nilThreshold$\leftarrow$ CalcNilThreshold(regularTakes)\;
    \eIf{expNilScore $>$ nilThreshold}{
       Return 0\;}{ Return regularTakes\;}
    \caption{\BIS algorithm}\label{alg:G bidding}
    
\end{algorithm}

 The \textit{regularTakes} variable is an estimation of the  number of tricks that the \BIS agent can take with high probability. This is a rule-based heuristic estimation based on the current hand and the precomputed probability tables. Details in Sec.~\ref{sub:regular_bid}.
 Similarly, the \emph{nilValue} (line~3) is a heuristic rule-based estimation of the chances to succeed in a nil bid (Sec.~\ref{subsec:nilvalue}). Since the accuracy of this estimation is critical,  we used data from online games to generate a more accurate probability estimation  \emph{nilProb}.  
 Generating the Success Curves used in line~4 is the main innovative part of the algorithm. The computation and use of success curves is detailed in Sec.~\ref{subsec:nil_threshold}.

The \emph{nilThreshold} in the deployed version of \BIS is a constant value (typically 25 points, but may be higher or lower under endgame conditions that we will later explain). We also consider a more structured way to compute the threshold.

\subsection{Regular bid}\label{sub:regular_bid}
In a regular bid, \BIS{} tries to estimate the largest number of tricks it can take with high probability.
Five features are considered to calculate the regular bid of a hand: (1) side suits high cards (2) spades high cards, (3) long spades suit, (4) short side suits accompanied with spades and (5) the sum of the previous bids.  Features 2,3 and 4 are not completely disjoint: the value of a spade card is the maximum between the value it gets from high/long spades and the value it gets from being a potential cut at a short side-suits. 

\subsubsection{Side Suit High Cards}  
We use a simplifying assumption that neglects the probability of \textit{finesse}\footnote{A finesse is a method of playing your cards to win a trick with a card lower than your opponents highest card.}. Thus the first, second and third tricks of each suit will be won by either the $A,K,Q$, respectively, or by a spade cut. 
Table~\ref{table:side suit high card's value vs 2 opponents}  presents the probability that in a specific suit, no opponent is void, singleton or doubleton (columns), given the number of cards the agent holds from that suit (rows).  These are exactly the probabilities that the agent will take a trick with the $A,K$ and $Q$ respectively (when neglecting the probability of finesse, and having enough blockers).  Values in parentheses are worth 0 since the high card does not have enough blockers to be played at the given trick. 

For example, in the hand agent holds at Fig.~\ref{fig:table}, the value of the $K\clubsuit$ is $0.678$. This is the probability that both opponents will not be able to cut the second $\clubsuit$ trick, given the agent holds five $\clubsuit$ cards (marked with $*$ in Table~\ref{table:side suit high card's value vs 2 opponents}).   
The value of $Q\diamondsuit$ is $0$ (marked with $**$) since agent does not have enough blockers to be played on the third trick.

When an opponent has bid nil or is known to be void at spades, only one opponent may cut so the probabilities are different, those probabilities are presented at Appendix \ref{app:Side-suit-high-cards-for-diffrenet-number-of-opponents}.

\begin{table}
\begin{center}
\caption{Side suit high card's value.  Calculated as the probability that the first/ second/ third trick can not be cut by an opponent, taking into account the player's number of cards from that suit.   
}
\label{table:side suit high card's value vs 2 opponents}
\resizebox{0.9\columnwidth}{!}
{
    \begin{tabular}{|c|l|l|l|}
    \hline
     &\multicolumn{3}{|c|}{Probability that both opponents have:}\\
    \multicolumn{1}{|l|}{\begin{tabular}[c]{@{}l@{}}cards in \\ side-suit\end{tabular}} & \textbf{\begin{tabular}[c]{@{}l@{}} $>0$ cards\end{tabular}} & \textbf{\begin{tabular}[c]{@{}l@{}} $>1$ cards \end{tabular}} & \textbf{\begin{tabular}[c]{@{}l@{}} $>2$ cards\end{tabular}} \\ \hline
    0 & (0.997) & (0.966) & (0.817)  \\ \hline
    1 & 0.994 & (0.942)~*** & (0.733)~**\\ \hline
    2 & 0.990 & 0.907 & (0.624)\\ \hline
    3 & 0.983 & 0.855 & 0.489\\ \hline
    4 & 0.970 & 0.779 & 0.350\\ \hline
    5 & 0.948 & 0.678~* & 0.212\\ \hline
    6 & 0.915 & 0.544 & 0.095\\ \hline
    7 & 0.857 & 0.381 & 0.025\\ \hline
    8 & 0.774 & 0.214 & 0\\ \hline
    9 & 0.646 & 0.074 & 0\\ \hline
    10& 0.462 & 0 & 0\\ \hline
    11& 0.227 & 0 & 0\\ \hline
    \end{tabular}
}
\end{center}
\end{table}
\vspace{-6mm}
\subsubsection{Spades High Cards} 
The $A\spadesuit$,$K\spadesuit$,$Q\spadesuit$,$J\spadesuit$ are each worth one trick if they are \textit{mostly protected}.  A spade high card is mostly protected if it has more spades than the number of un-owned higher rank spades. This notion comes from the blog \emph{Tactics and Trickery} by Tyler Wong~\cite{TylersSpadesblog}.  Formally, the $A\spadesuit$ is worth one trick.  The $K\spadesuit$ is worth one trick if the hand contains another spade.  The $Q\spadesuit$ ($ J\spadesuit $) needs 2 (3) protectors in order to be counted as a take. 

\subsubsection{Spades Long Suit} 
Every spade beyond the forth is counted as a take since given the agent holds five spades or more, it is likely no opponent is holding five spades.  
For example, the spades at Fig.~\ref{fig:table} ($ A\spadesuit,K\spadesuit,J\spadesuit,6\spadesuit,2\spadesuit $) are worth 4 takes: two takes for the $ A\spadesuit,K\spadesuit $, the $J\spadesuit$ together with the $2\spadesuit$ as a blocker is worth another take, and another take is due to the fifth $\spadesuit$.

\subsubsection{Short Side Suits with Uncounted Spades} 
\BIS's value of short side suits is the probability that it is the only player that can cut in the specific trick, (otherwise, opponents might over cut).  Table \ref{table:side suit high card's value vs 2 opponents} show this probabilities for void, singleton and doubleton.  For example, the cutting value of a Singleton $\diamondsuit$ and two unassigned spades, is the probability that no opponent is having the possibility to overcut on the second and third $\diamondsuit$ tricks, that is 0.942 + 0.733 = 1.675 (marked with ***,** in Table~\ref{table:side suit high card's value vs 2 opponents}).  Each $\spadesuit$ card is counted once where it produce the highest value, either for high/long spades or for cutting short side-suit.

\medskip
In total, in the example hand displayed at Fig.~\ref{fig:table}:
\begin{itemize}
\item The side suit high cards contribute $0.678$ (for $K\clubsuit$) and $0.99$ (for $A\heartsuit$). While the $Q\heartsuit, Q\diamondsuit$ contribute 0. 
\item The $A\spadesuit,K\spadesuit$ contribute 1 take each.
\item  The $J\spadesuit,6\spadesuit,2\spadesuit$  can be counted as a high card ($J\spadesuit$) with a blocker and a fifth $\spadesuit$, and contribute $1+1=2$.
\item Alternatively, the $J\spadesuit,6\spadesuit,2\spadesuit$, can be counted as short side suits cuts,  which worth $0.942 + 0.733 + 0.624 = 2.3$.
\end{itemize}
Thus the expected amount of regular takes is  $0.678+0.99+2+\max\{2,2.3\}=5.89$. To this value we add a factor which determined by the sum of previous bids and then it is rounded to the nearest integer to determine the value of regularTakes in line~2 of Alg.~\ref{alg:G bidding}.

\subsection{Nil-Value}\label{subsec:nilvalue}
\BIS heuristiclly estimate the probability of a successful nil bid, we denote this estimation as the hand's nilValue. 

Our main relaxation is \emph{almost-suit-independence}:\footnote{Suit-independence occurs since players must follow the leading suit if they can, it breaks down when a player is void in the leading suit.} the probability of taking 0 cards from a given suit, depends only on the cards of that suit.  Formally:

$$ Pr(nil|hand) \approx \prod_{suit \in\{\clubsuit,\diamondsuit,\heartsuit,\spadesuit \} } Pr(nil(suit) | hand \cap suit), $$

This relaxation reduces the number of unknown location of cards from 39 to 13 (minus our holdings from that suit).  Thus, enables to efficiently simulate every single suit set of cards in the agent's hand.
This relaxation was used in \cite{buro2009skat,  edelkamp2018challenging} to evaluate nil winning probability in a resembling trick-taking game named Skat. 

Therefore, the problem reduces to estimating $$Pr(nil(suit)|hand \cap suit)$$ in a given suit.  Our second simplifying assumption is that in each suit, cards beyond the three lowest are not dangerous.\footnote{\label{note1}The fourth card in a suit is seldom dangerous since when holding 4 cards from the same suit, only $8\%$ of the deals will allow the opponents to lead a forth trick while the partner can not cover by cutting.  
}

\paragraph{Monte Carlo deals}
As a preprocess, we evaluated the probability of taking zero tricks with each possible set of a single suited cards.  
 The event ``$nil(suit)$" depends not only on dealt cards, but also on how players will play, which makes the evaluation ambiguous.  Therefore we evaluate a different event $cnil(suit)$ (`cards nil') which only depends on the cards players have. Formally, this event means that ``on all tricks of the relevant suit, both opponents can play under one of the agent's cards, and  partner cannot cover that card".\footnote{This event leads to failed nil under the following assumptions: (1) no cards from this suit are played on tricks where different suit was lead, (2) both opponents are playing under the agent's card if it is the highest on the table, otherwise they play their highest, (3) partner covers with high card when able, (4) partner can always cut when she is void in the suit.}  

We  evaluate the probability of the complement event $Pr(\neg cnil(suit)|hand\cap suit)$. There are four cases depending on how many cards are in the suit, and each of them can be written as a union of simple events:
\begin{enumerate}
    \item In a \textbf{void suit} $\emptyset$ the nil can never be set, so $Pr(\neg nil(suit)|hand \cap suit)=0$ (note that in this case there is no difference between $nil$ and $cnil$). 
     
    \item {in a \textbf{singleton suit} $\{x\}$:}  both opponents are either void or have at least one card smaller than $x$, and partner isn't void and has no higher card.
    
    \item {in a \textbf{doubleton suit} $\{x\ol x\}$ is the union of the following two events:}
        \begin{enumerate}
        \item {set at the smallest card $x$:}  same as in the singleton suit case. 
        \item {set at the second smallest card $\ol x$:}  both opponents are singleton/void or have at least two cards smaller than $\ol x$, and partner isn't singleton/void or has no two higher cards than $\ol x$.
       \end{enumerate}  
        
    \item {in a \textbf{suit with three cards or more}\footnoteref{note1} $\{x\ol x \oll x\}$ is the union of the following three events:} 
        \begin{enumerate}
            \item {set at the two smallest cards $x$ or $\ol x$:} same as in the doubleton case. 
           \item {set at the third card $\oll x$:}  both opponents are doubleton/ singleton/ void or have at least three cards smaller than $\oll x$, and partner isn't doubleton/ singleton/ void and has no higher card than $\oll x$.
        \end{enumerate}  
        
    \item[\textbullet] evaluating the $\spadesuit$ suit is the same except that partner can not cover by cutting and that a forth spade denials nil (which is a popular heuristics~\cite{cardgamesIO}).
\end{enumerate}
 
To evaluate the simple events used in the above cases, we made a table that contains an entry for each outcome. This is similar to Table~\ref{table:side suit high card's value vs 2 opponents} but with many more rows, a row for each possible set of cards from the suit.  For each entry, we uniformly deal the rest of the suit between the other players 100K times.

\begin{figure*}[t!]
    \centering
    \includegraphics[width = 430pt]{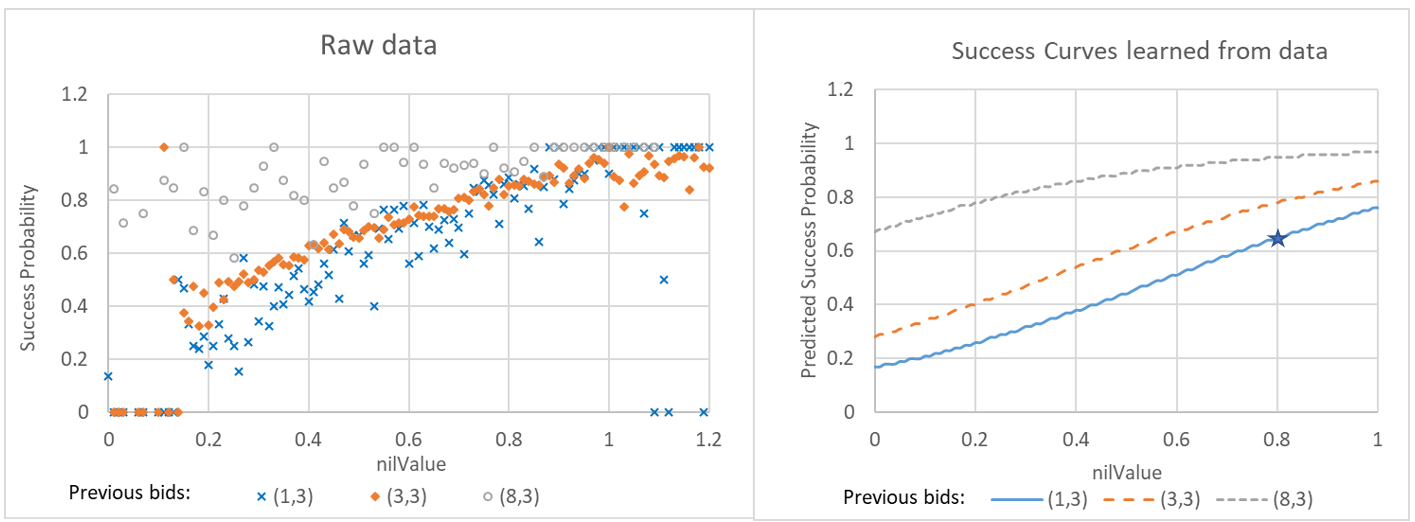}
    \caption{Left: the nilValue and the actual nil success probability of the agent when bidding third, after the RHO bid $3$, and the partner bid $1$, $3$, or $8$.  Right: the learned success curves for the three bidding sequences on from the left figure.}
    \label{fig:SC}
\end{figure*}

For example, the nilValue of the hand in Fig.~\ref{fig:table} is calculated as follows:
\begin{itemize}
  \item $\clubsuit: K,9,5,4,3$. The 3 lowest cards are perfectly safe, thus  $Pr(cnil(\clubsuit)|hand)= 1$.
  \item $\diamondsuit: Q\diamondsuit$.  The singleton $Q$ has $57.8\%$ to be covered,  either when the partner is void, or when the partner holds either the $K$ or the $A$, or when one of the opponents  holds only higher $\diamondsuit$. Thus $Pr(cnil(\diamondsuit)|hand)= 0.578$.
  \item $\heartsuit: A,Q$.
  According to the respective doubleton  table, $Pr(cnil(\heartsuit)|hand) < 0.001$. It is non-zero only due to the chance is that partner is void. 
  \item $\spadesuit: A,K,J,6,2$. The $A$ guarantees a failed nil, thus $Pr(cnil(\spadesuit)|hand)=0$.
  
  \item Thus, the Nil-Value is $1\cdot0.578\cdot 0.001 \cdot 0 = 0$
\end{itemize}

When the hand contains a void suite, we multiplie the nilValue by a constant factor of $1.15$  (hence `almost-independence'), which is a reasonable approximation that does not require to recompute all the tables. 

\subsection{Supervised Learning of the Nil Success Probability}\label{subsec:nil_threshold}
The nilValue computed in Sec.~\ref{subsec:nilvalue} is providing us with some estimation of the actual probability to succeed in a nil bid. However, we can get a more accurate evaluation of the probability if we take into account the bids of the previous players in this round.

To estimate the nil success rate, we used data from games of the earlier version of \BIS that does not use the learning component, combined with the playing module \SRP (see Section~\ref{subsec:Competition with other bots}).  
We extracted  2 million rounds from online games played during December~2018 (all games had three bots and a single human player). We only used rounds where the agent bids nil.

For every sequence of previous bids and any nil value, we counted how many nil bids were successful. See Fig.~\ref{fig:SC}(left) for three such sequences. We can see that the nilValue is indeed positively related to the actual success probability, but is not in itself a good approximation, as the probability highly depends on the previous bids.  
Since \BIS  never bids nil in some situations (e.g. low nilValue), we also used a noisy variant of \BIS for better exploration. 


Ideally, we would get a curve of $Pr(nil | nilValue)$ for every possible bid sequence.   However, there are $1+14+14^2+14^3= 2955$ bidding sequences (0-3 previous bidders, each bid in the range of 0-13). Some of them have enough data to provide a good estimation, but for other bid sequences data is scarce. E.g., there are over 20K rounds for the previous bid sequence ``(3,3)", but only about 700 rounds for the sequence ``(8,3)", which means no data at all for some nilValues.  

 To generate the success curves for all bidding sequences, we trained a binary Logistic Regression model on our collected data. The model estimates the nil success probability for each of the $2955$ bidding sequences and every possible nilValue. 
 
  We then utilize our trained model to retrieve a nil success probability estimate for all possible bidding sequences and nilValues, including for values that do not occur in our data. See Fig.~\ref{fig:SC}(right) for the learned success curves of the three sequences mentioned above.

We generated all $2955$ success curves offline, and stored them as the $SC$ tables. The bidding algorithm (see row~4 of Alg.~\ref{alg:G bidding}) uses the relevant success curve for the actual previous bids, and returns the estimated nil success probability as $SC(PrevBids,nilValue)$.  For example, if the agent bids third, previous bids are $(1,3)$,  and the calculated nil value is $0.8$ then the estimated nil success probability is $65\%$ (marked with a star in Fig.~\ref{fig:SC}.

Besides Logistic Regression, we experimented with neural networks, random forests and linear regression as well, and got similar results, thus we chose to use Logistic Regression for the following reasons:
\begin{itemize}
    \item Interpretable - we can easily understand how it weights each feature, as opposed to neural networks.
    \item Easy to implement and train.
    \item Explicitly models the probability estimates we are interested in.
    \item It produces a probability that is monotone in the nil value, as opposed to other methods.
\end{itemize}


\subsection{End-of-game Bidding Modifications}\label{subsec:endgame}
In most rounds, maximizing the expected points in the round is a good approximation to maximizing the winning probability in the game.  However when at least one partnership will win the game by fulfilling their contract those two objective differ widely.  \BIS becomes risk seeking when opponents are about to win and risk averse when partnership is winning.  
An example for such modification is the `complete to 14' bid which means betting on the opponents to fail their bid (as there are only 13 tricks).  While this modification may yield negative expected points, it increases the winning probability when opponents are close to winning the game. Those heuristics are detailed at Appendix~\ref{subsub:End-of-game}. 

\section{Experimental Evaluation}
In this section, we evaluated our bidding algorithm against other bidding algorithms, and against human players.
We then  evaluate the impact of the different components of our bidding algorithm. 

\subsection{Competition with other bots}\label{subsec:Competition with other bots}
\vspace{-3mm}

\paragraph{Setting} We matched \BIS with three competing rule-based bidding algorithms (see below). In every game there was one \BIS partnership (two agents using \BIS for bidding) and one competing partnership. To ensure that only the bidding component is evaluated, all four players used the same playing module  (see below).

In each comparison 10K games where played, which is about 35K rounds. Unless stated otherwise, we used a winning goal of 200 points and losing goal of -100 points.\footnote{The (+200/-100) goal was a common setting on the application at the time. We received similar results with other goals.}

\noindent
\textbf{Competing Algorithms:}
\begin{description}
    
    \item[IO]  This agent is implemented on \url{Cardsgame.io} website  and uses a fairly simple bidding method.  It uses the following nil classifier:  if the regular bid $\leq 3$ and partner's bid $\ge 4$ and hand contains no A or K, and no A-T of spades and no more than 3 spades, then bid nil.  
    It uses the following regular bid:  each high spade (A to T) is worth 1 trick, each low spade (9 to 2) is worth 0.4 trick, in side suits, each A is worth 1 trick, each $\{AK\}$ is worth 2 tricks and each $\{Kx\}$ is worth 0.5 a trick.
    We did not get access to the playing module of this agent.
    
    \item[MS]  This is an implementation of the bidding scheme denoted as `simple bidding' which appears in the book  \emph{Master Spades}~\cite{Spades2002Strategy}. 
    The instructions in the book for bidding nil (and also the playing instructions) are not concrete enough to write them down as an algorithm.
    We therefore combined the bidding algorithm with the na\"ive nil classifier of \RB.    The regular bid is the following: Nine cards are worth one trick each: Aces, non-singleton Kings and the $Q\spadesuit$ (if it is not a singleton or doubleton that does not contain the $A\spadesuit$).  Each $\spadesuit$ beyond the first three is worth a trick while a void or a singleton spade reduce the bid by one.  A side-suit void or singleton, together with exactly three spades increases the bid by one.
    
    \item[RB] Rule-based Bidder.  This is the previous bidder that was implemented in the application we use for evaluation.  Its regular bid calculated as a sum of values of each card in  hand.  Each card has a value that depends on the number of cards from that suit in the hand.  The na\"ive nil classifier bids nil if the lowest, second lowest and third lowest cards from each suit are not larger than $5,8$ and $10$ thresholds respectably.  
\end{description}

\textbf{Playing modules:}
\begin{description}
    \item[WRP]  A Weak Rule-based Player. Strength of an average recreational human player. When combined with the RB bidding module, wins almost $50\%$ of the games when plays vs. recreational human players.
    \item[SRP]  A Strong Rule-based Player - When combined with the RB bidding module, wins $56\%$ of games vs. RB+\WRP; and $54\%$ of the games vs. recreational human players.
    \item[UCT] An implementation of the \UCT algorithm~\cite{kocsis2006bandit} for the game of Spades. We use a time limit of 3 seconds to decide the number of samples.   Further details about our implementation and the limitations of \UCT are found at Appendix~\ref{app:uct}.
\end{description}

\noindent
We can also use the general \UCT algorithm for bidding and not just for playing. We denote by \UCT(X) the SRP playing module where the last X tricks are replaced with \UCT. 


\paragraph{Results}
Table \ref{table:results_sim} shows that for three different playing modules, the bidding module of \BIS is stronger than the other bidding modules. 

\begin{table}
\begin{center}
{\caption{Comparison of \BIS bidding to other bidding algorithms.  In each comparison, the playing module is fixed for both bidding modules.  The comparison is in two aspects: win ratio and average points per round.}\label{table:results_sim}}
\resizebox{\columnwidth}{!}
{
    \begin{tabular}{|l|l|l|l|}
    \hline
    \textbf{\begin{tabular}[c]{@{}l@{}}Opponents' \\ Bidding\end{tabular}} & \textbf{\begin{tabular}[c]{@{}l@{}}Playing \\ Module\end{tabular} } & \begin{tabular}[c]{@{}l@{}}\BIS's \\ win rate\end{tabular} & \textbf{\begin{tabular}[c]{@{}l@{}}Average points \\\BIS{} : opponents \end{tabular}}  \\ \hline
    \RB & \WRP    & 51.9\% & \textbf{53.2} : 50.0 \\ \hline
    \MS & \WRP    & 66.9\% & \textbf{50.1} : 38.2 \\ \hline
    \IO & \WRP    & 68.6\% & \textbf{52.8} : 46.5 \\ \hline
    \RB & \SRP    & 52.3\% & \textbf{56.2} : 52.8 \\ \hline
    \MS & \SRP    & 67.7\% & \textbf{50.8} : 38.4 \\ \hline
    \IO & \SRP    & 67.1\% & \textbf{51.6} : 46.3 \\ \hline
    \RB & \UCT(3) & 52.1\% & \textbf{52.0} : 48.5 \\ \hline
    \RB & \UCT(5) & 52.9\% & \textbf{48.5} : 40.3 \\ \hline
    \end{tabular}%
}
\end{center}
\end{table}


Our results shows that the \UCT playing module is stronger than \SRP only when activated at the last several tricks, this is because of the three seconds time limit which does not allow the \UCT to search accurately when activated earlier.  If we would like to use \UCT as a bidding module, it would have to search a huge space.  \rmr{removed:By extrapolation we can safely assume that using \UCT for the bid, while having a time cap of 3 seconds is inferior to \BIS.}
We are currently looking into how \UCT can be combined with our \BIS agent without using excessive computation time.




\subsection{Playing against humans}
\paragraph{Setting and dataset}
Our \BIS agent was deployed in a popular mobile Spades application.\footnote{We disclose the name of the platform in a note to the reviewers.} We extracted more than 400K rounds from games played during October 2019,  between a partnership of two BIS+\SRP agents and a partnership composed of a BIS+\SRP and a human player.  


\paragraph{Results}
 The overall winning rate of the \BIS partnership is $56\%$ of all games, which are 2 percentage points above the performance of the previous rule-based partnership used by the platform (\RB bidding module with the same \SRP playing module). Note that this data suffers from selection bias because games were only recorded if they ended, while humans tend to quit games when they are losing badly.

 To better understand the strengths and weaknesses of \BIS we divided all rounds to types according to the bids made by players. 

\begin{figure}
    \centerline{ \includegraphics[width=0.9\linewidth]{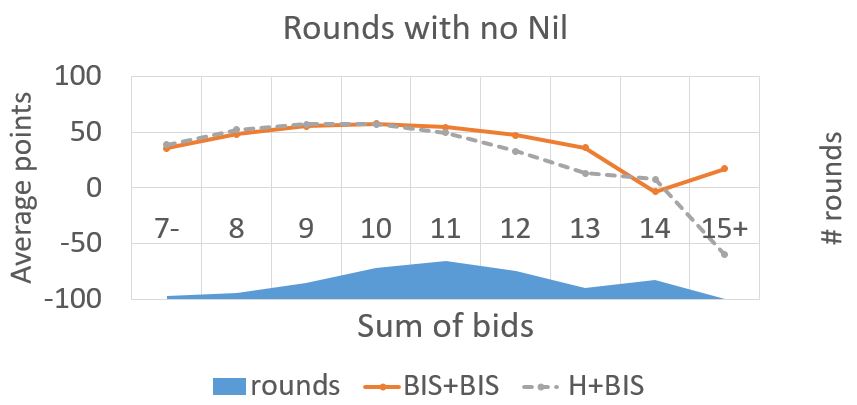} }
    \caption{Average points per round in rounds where no player has bid nil, broken down based on the sum of the four bids.} \label{fig:Average_points}
\end{figure}

 Table~\ref{table:results_real} shows  that in almost all round types, the BIS+\SRP partnership obtains a higher score.
 This is true also when partitioning no-nil rounds according to the sum of bids (Fig.~\ref{fig:Average_points}).
 The biggest points gaps in favor of BIS+\SRP are in `double nil' rounds (both partners bid nil) and when the total bids exceed $14$.  We conjecture that those rounds are a result of bidding blunders made by humans.\footnote{Double nil bid is rarely beneficial (\BIS never bids nil if its partner bid nil).  Sum of bids higher than 14 is usually an indication that a human player is vastly overvaluing their hand, since \BIS bids conservatively. }

One exception where the BIS+H partnership scores higher is when bids sum up to exactly $14$. This is explained by \BIS end-of-game bidding modifications (see Sec.~\ref{subsec:endgame}).
The other exception is when the human's partner bids Nil, demonstrating that the poor performance of the partnership is due to the human part. 


\begin{table}
\begin{center}
\caption{Comparison between a partnership of two BIS+\SRP against BIS+\SRP and a human.  The comparison is broken down by round types.  The BIS+\SRP partnership is generating more points in rounds containing nils.} \label{table:results_real}
\resizebox{\linewidth}{!}
{
    \begin{tabular}{|l|l|ccc|ccc|}
\hline
nil position & $\#$ rounds & \multicolumn{3}{|c|}{Average points} & \multicolumn{3}{|c|}{Successful nils}\\ 
             &        & BIS+BIS &:& H+BIS &  BIS+BIS& :& H+BIS \\ \hline
BIS+BIS      & 117782 & 57.3 &:& 47.7 & 68.8\% & :& -   \\ \hline
H            & 73157  & 50.7 &:& 49.2 & -&:& 63.9\%  \\ \hline
Partner of H    & 57392  & 51.9 &:& 62.2 & -& :& 70.0\% \\ \hline
Nil vs. Nil   & 102132 & 67.4 &:& 60.5 & 44.2\% & :& 39.6\% \\ \hline
Double Nil   & 691    & 64.5 &:& -29.3 & - &:& 32.4\% \\ \hline
No Nil       & 95928  &	48.9 &:& 43.6 &   -&:&- \\ \hline
\end{tabular}
}
\end{center}
\end{table}

\subsection{Impact of \BIS components}
This section aims to reveal the significance of several components in \BIS bidding by comparing a \BIS partnership to a partnership that has the specific component disabled or modified. In all simulations, all four bots use the \SRP player.  

\begin{table}
\begin{center}
    \caption{\BIS Vs \BIS with a single component disabled} \label{table:impact_of_components}
    \begin{tabular}{|l|l|l|}
    \hline
    Opponents & \begin{tabular}[c]{@{}l@{}}BIS's\\ win rate\end{tabular} & \begin{tabular}[c]{@{}l@{}}Average points\\ BIS : opponents\end{tabular} \\ \hline
    \begin{tabular}[c]{@{}l@{}}Single success curve\end{tabular} & 51.2\% & 44.7 : 43.6 \\ \hline
    \begin{tabular}[c]{@{}l@{}}No end-game conditions,\\ winning goal of 200\end{tabular} & 52\% & 48.3 : 49.9 \\ \hline
    \begin{tabular}[c]{@{}l@{}}No end-game conditions,\\ winning goal of 1\end{tabular}   & 52.8\% & 43.5 : 52.9 \\ \hline
    \end{tabular}
\end{center}
\end{table}

\subsubsection{Success Curves}
A simple variant of the \BIS bidding algorithm would use a single Success Curve, without taking the previous bids into account. The current \BIS algorithm beats the simple variant in $51.2\%$ of the games, and gains $2.5\%$ higher score on average. This means that the success curves are responsible for about $1/4$ of the overall improvement that \BIS obtains over the best rule-based bidder \RB.


\subsubsection{End-of-game Bidding Modifications}
 The end game conditions are a set of heuristics described at Subsection \ref{subsec:endgame}.  \BIS{} with the end game conditions won $52\%$ of the games under the usual winning goal of 200 points. As expected, \BIS with the endgame module obtains \emph{fewer} points per round.  When the winning goal was set to 1 point (single round games), the win rate increased while the average points decreased.

\subsubsection{Bid-sensitive Nil Threshold}
The \BIS algorithm uses a threshold as a cutoff point to decide whether to bid nil (see Lines~6-7 in Alg.~\ref{alg:G bidding}). The current algorithm uses a constant threshold (specifically, 25) that is based on the expected score of  a non-nil bid.   It may seem wiser to use more available information to determine this threshold. 

Indeed, we implemented a variation of the \BIS algorithm (BIS$^*$), which tries to evaluate the expected score \emph{of the partnership} once if the agent would bid nil and second if she will bid a non-nil bid, using the regularTakes value and the bid of the partner, if known.

When playing against each other, BIS$^*$ was slightly worse. One possible explanation is that an inaccurate estimation is worse than using a constant threshold. We hope to better understand the weak points of the estimation and improve the threshold decision in future work.  

\section{Conclusion}
This work is the first to publish a Spades bidding algorithm that outperforms recreational humans.  Our hope is that it will serve as a baseline for future work which will allow other teams to build stronger Spades bidding modules.

\BIS is flexible in the sense that it can bid in several variations of Spades, such as Cutthroat Spades (i.e. Solo), Mirror, Spades with jokers, Whiz and Suicide, each of them require very little game-specific modifications.
We conjecture that the methods we used might produce strong bidding modules in other trick-taking games such as Skat, Whist and Callbreak.  In the first two games, a nil classifier is a major part of the bidding and our nil classifier, with slight modifications can be used.  Our regular bid evaluator needs slight modification to the value of cards in order to be used in those other games.

The main takeaway message that goes beyond applications to trick taking card games, is our approach of combining rule-based heuristics and learning. That is, we first generate a rule-based heuristics (in this case, of the probability to succeed in a nil bid), and then apply machine learning on past data, using this heuristic as the main feature, to get an improved estimation. 

Future research is focused on better tackling the weak points of the bidding module (cases where probability estimations are off), and on improving the playing module. Our goal is to develop a combined Spades agent with super-human  strength.     


\ack We would like to express our gratitude to Nathan Sturtevant and Stephen Smith for their experts advice, many thanks to Einar Egilsson for granting us access to his bidding algorithm.\footnote{Available to play at cardgames.io/spades/}  We would like to thank an anonymous game studio that gave us access to their data of completed games which contribute to our evaluation of \BIS performance.

\bibliographystyle{plain}

\clearpage

\appendix


\section{Side-suit high cards value for different number of dangerous opponents} \label{app:Side-suit-high-cards-for-diffrenet-number-of-opponents}
When evaluating the value of side-suit high cards, the following couple of charts complete the picture presented in Table~\ref{table:side suit high card's value vs 2 opponents} for the cases where the number of opponents that might cut your high cards changes from the default `2' to `3' or `1'.  This change happens in two cases: firstly, when an opponent bids nil and is not been set yet, (she will tend to avoid cutting in order to protect her nil).   Secondly, when playing the Cutthroat variant where there are no partnerships, the player is facing three opponents that might cut her high cards.  \BIS uses the probability that on the first, second or third trick lead by a given suit, the opponents will still hold cards from that leading suit, thus will be unable to cut.  This Monte-Carlo estimation takes into account the the number of cards the agent holds from that suit and the number of opponents that might cut.

\begin{table}
\begin{center}
\caption{Side suit high card's value - 1 opponent that might cut.  }
\label{table:side suit high card's value vs 1 opponents}
\resizebox{\columnwidth}{!}
{
    \begin{tabular}{|c|l|l|l|}
    \hline
     &\multicolumn{3}{|c|}{Probability that the opponent has:}\\
    \multicolumn{1}{|l|}{\begin{tabular}[c]{@{}l@{}}cards in \\ 
    side-suit\end{tabular}} & \textbf{\begin{tabular}[c]{@{}l@{}} $>0$ cards\end{tabular}} & \textbf{\begin{tabular}[c]{@{}l@{}} $>1$ cards \end{tabular}} & \textbf{\begin{tabular}[c]{@{}l@{}} $>2$ cards\end{tabular}} \\ \hline
    0 & (0.998) & (0.983) & (0.910) \\ \hline
    1 & 0.997 & (0.971)  & (0.866) \\ \hline
    2 & 0.994 & 0.954 & (0.809) \\ \hline
    3 & 0.992 & 0.927 & 0.733   \\ \hline
    4 & 0.985 & 0.891 & 0.648  \\ \hline
    5 & 0.974 & 0.835  & 0.546  \\ \hline
    6 & 0.957  & 0.761   & 0.426  \\ \hline
    7 & 0.928  & 0.667   & 0.308   \\ \hline
    8 & 0.886  & 0.546 & 0.195   \\ \hline
    9 & 0.819  & 0.410   & 0.100   \\ \hline
    10 & 0.715 & 0.252   & 0.030  \\ \hline
    11 & 0.561 & 0.106  & 0         \\ \hline
    12 & 0.336 & 0 & 0         \\ \hline
    \end{tabular}
}
\end{center}
\end{table}

\begin{table}
\begin{center}
\caption{Side suit high card's value - 3 opponents that might cut, used in the Cutthroat variant.}
\label{table:side suit high card's value vs 3 opponents}
\resizebox{\linewidth}{!}
{
    \begin{tabular}{|c|l|l|l|}
    \hline
     &\multicolumn{3}{|c|}{Probability that the three opponents have:}\\
    \multicolumn{1}{|l|}{\begin{tabular}[c]{@{}l@{}}cards in \\ 
    side-suit\end{tabular}} & \textbf{\begin{tabular}[c]{@{}l@{}} $>0$ cards\end{tabular}} & \textbf{\begin{tabular}[c]{@{}l@{}} $>1$ cards \end{tabular}} & \textbf{\begin{tabular}[c]{@{}l@{}} $>2$ cards\end{tabular}} \\ \hline
    0 & (0.996) & (0.949) & (0.729) \\ \hline
    1 & 0.992 & (0.915) & (0.605) \\ \hline
    2 & 0.986 & 0.862 & (0.450) \\ \hline
    3 & 0.974 & 0.784 & 0.275 \\ \hline
    4 & 0.955 & 0.672 & 0.110 \\ \hline
    5 & 0.924 & 0.523 & 0 \\ \hline
    6 & 0.872 & 0.338 & 0 \\ \hline
    7 & 0.790 & 0.145 & 0 \\ \hline
    8 & 0.664 & 0 & 0 \\ \hline
    9 & 0.480 & 0 & 0 \\ \hline
    10 & 0.240& 0 & 0 \\ \hline
    \end{tabular}
}
\end{center}
\end{table}

\section{Bidding patterns}
\begin{figure}
\centerline{ \includegraphics[width = 300pt, height = 150pt, keepaspectratio]{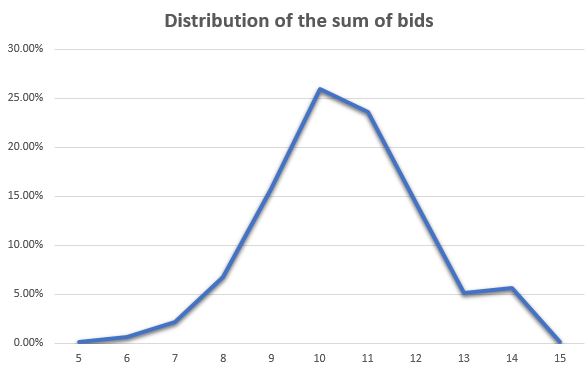} }
\caption{The sum of four players bids.  The average value is 10.55.} \label{fig:sum_of_bids}	
\end{figure}

Figure~\ref{fig:sum_of_bids} shows the sum of all four players' bids in a round.  As measured from a dataset of millions of games.  Note that Spades scoring rules punish severely a set (failed) contract, while the punish for overtakes is relatively small.  This is the reason that although there are 13 tricks in a round, the average sum of bids is about 10.55.

\section{Branching factor}
Spades playing phase has a low branching factor. With an average number of possible actions of approximately $3.6$.  Note that the branching factor of the bidding is 14.

\begin{figure}
    \centerline{
      \includegraphics[width = 300pt, height = 150pt, keepaspectratio]{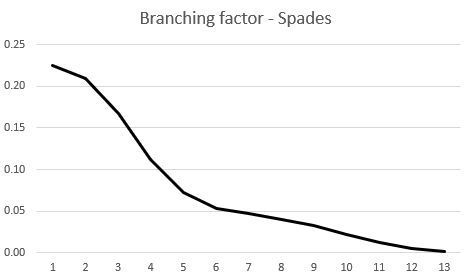} 
    }
    \caption{The number of possible actions in the playing phase of Spades.  Note that in the bidding phase each player is face with a single 14-options decision. } \label{fig:Branching_factor}	
\end{figure}

\section{Interesting aspects in the playing algorithm \SRP{}} \label{sec: playing}
Since \SRP{} is a knowledge based algorithm, it is impossible to described it briefly, yet few sub-functions are interesting and can be beneficial in other games as well.

\subsection{Number of sure future takes}  
This function determines the number of guaranteed takes a hand possess. This is a worst-case analysis.  This analysis is equivalent to the use of the ``normal strategy" done by Johan Wastlund at \cite{wastlund2005two}. Computing the sure winners is equivalent to the toy game `1 suit, 2 players Whist', this is true because in the worst case, all unseen spades are held by a single opponent and all the spades tricks are lead by the agent.    Only Spades cards can be sure winners, side suit cards cannot be a guaranteed winners since a side suit might never be lead.  Spades can be sure winners by two reasons: either by being the boss or by having more spades than the unseen spades.  For example, $A\spadesuit$ is a sure take, $\{K\spadesuit, Q\spadesuit\}$ is a single sure take, holding 4 spades cards when 7 spades cards were played is two sure takes.   

Knowing how many sure takes a hand possess before those cards are played is beneficial in the following scenarios
\begin{itemize}
    \item \textbf{Avoiding bags} - if agent knows that the partnership's bid is guarantee and setting opponents has a low chance, she can avoid taking extra tricks.
    \item \textbf{Cover niling partner} - if the agent knows her bid is guaranteed, she can cover her partner with high cards that are not boss,  without having to save them in order to make her own contract.
    \item \textbf{Setting niling opponent} - agent can play to set niler without saving risking her own contract.
\end{itemize}

\subsection{Goal-choosing}
Unlike Bridge, were an agent should always maximize the tricks her partnership takes, in Spades there are situations where avoiding tricks is the best action.  Examples for goals dilemmas:
\begin{itemize}
    \item In rounds with a nil bid by opponent there is a conflict between two strategies, trying to set the niler or trying to set the coverer.  Identify when to switch from trying to set the niler into trying to set the coverer is still a challenge.
    \begin{enumerate}
        \item Setting niler opponent - while if succeed the reward is huge, some times trying to set the niler results in easy takes for the coverer, which risk the partners' bid.
        \item Setting coverer opponent - since the coverer is focusing in playing high cards to protect her partner's nil, his own bid is often vulnerable.
        \item Avoiding bags - in situations where all bids are either met or sure to be set, the agent try to avoid taking additional bags.
    \end{enumerate}
    
    \item In rounds with no nil
    \begin{enumerate}
        \item Setting opponents bid.  Disadvantages: risk of getting unnecessary bags without setting the opponents.
        \item {Bagging opponents}.  Disadvantages: risk setting your own bid.
    \end{enumerate}
    
     \item In rounds with nil partner
    \begin{enumerate}
        \item A trade-of exists between covering your partner nil and securing your own bid. When it is obvious that partner's nil will be set, it is better play to stop covering partners' nil and focusing on getting her own bid.  This module is not implemented, \SRP never abandon covering partner's nil until it is set.  As stated in \cite{whitehouse2013integrating}, failing to cover partner's bid is the major reason for complains of players about their partners, thus on this issue we sacrificing some playing strength in order to have happier human players.
    \end{enumerate}
\end{itemize}

\subsection{Signaling conventions in the playing phase}  
\begin{itemize}
    \item \textbf{High-low doubleton}
    It is a known convention in Bridge.  Playing an unnecessary high card, then on a later trick, playing a lower card means a doubleton suit.  Formally, if on the first trick from a suit, the agent follow under a boss, then on the next trick from that suit, it uses a lower card, then it is a signal that the agent had only 2 cards from that suit, meaning that the next lead in this suit can be cut, thus asking partner to lead that suit.
\end{itemize}

\section{Rounds types and end-game conditions}
The bidding phase determine the type of round.  Different types of round makes player pursuit different goals.  For example, rounds with high sum of bids encourage players to take more tricks than they need to set the opponents, rounds with low sum of bids encourage players to avoid taking many bags.  Rounds with a nil bid place the players in a dilemma whether they should try to set the niler or her coverer. \\
\SRP distinguish between 10 round types and play different at each of them:
\begin{enumerate}
    \item strong under - no nils, sum of bids $ \le 8$.
    \item under - no nils, sum of bids $ \in\{8, 10\} $.
    \item over - no nils, sum of bids  $ \in\{11, 13\} $.
    \item 14 - no nils, sum of bids is exactly $ 14 $.
    \item strong over - no nils, sum of bids $ \ge 15$.
    \item we nil.
    \item partner nil.
    \item opponents nil - a single player from the opponents bid nil.
    \item nil vs. nil - each partnership has a player which bid nil.
    \item double nil - both opponents have bid nil.
\end{enumerate}

On top of the round type, in rounds where at least one partnership can win the game, \BIS alter its bid from maximizing the expected points gain, to maximizing the winning probability.

End-game conditions:
\begin{enumerate}
    \item partnership can win the game on this round.
    \item Opponents can win this game on this round.
\end{enumerate}
When the agent can win the game, \BIS bids more conservative, that is, if the partnership is winning by a larger than 10 points margin and isn't close to the bags limit, then it decrease the bid by one.

When the opponents can win the game, \BIS alter it's bid such that if both partnerships will make their bid, the agent will win. If opponents have bid nil, \BIS is subtracting a trick from her bid and will try to set the niler at all costs.


\section{Other bidding Agents}

\subsection{Master Spades bidding}
This is an implementation of the guidelines described at Master Spades book \cite{Spades2002Strategy}.  While this book is good in teaching humans to play Spades well, it uses simple guidelines that address only the common situations and relay on human reason to make proper adjustments, thus implementing those guidelines results in sub-human bidding algorithm.  The book does not specify a nil classifier, so we used a naive nil classifier that is simple enough to be described in a book and then followed by humans.

The \textbf{Master Spades regular bid:} There are nine cards which are worth one trick each: Aces, non singletons Kings and the $Q\spadesuit$ (if it is not a singleton or doubleton that does not contain the $A\spadesuit$).  Each $\spadesuit$ after the first three worth a trick while a void or a singleton spade reduce the bid by one.  A void or singleton in a side-suit together with exactly three spades increases the bid by one.

\subsection{Rule-based bidder \RB{}}
This rule-based bidder has the strength of an average recreational mobile player.  In games with three of those agents and a human in the forth position, the partnership containing two agents wins just a bit more than $50\%$ of the games.

\paragraph{Naive Nil Classifier}  A naive classifier bids nil if in each suit, the first, second and third lowest cards are no greater than the ranks of $X,Y,Z$.  Through trial and error $X,Y,Z$ where set to $5$, $8$ and $T$ respectively. In addition, it require that the hand contain no more than 3 spades and that the partner's bid is not nil.  It's main problem is that it does not combines the vulnerabilities from the different suits.  While it is safe to bid nil with a hand that contain one vulnerable suit, it is not safe to bid nil when all the suits are vulnerable.  For example, the naive classifier will classify a hand like $2\clubsuit, 3\clubsuit, J\clubsuit, 2\diamondsuit, 3\diamondsuit, 4\diamondsuit, 5\diamondsuit, 6\diamondsuit, 7\diamondsuit, 8\diamondsuit, 9\diamondsuit, T\diamondsuit, J\diamondsuit$, as a no-nil hand, because of the venerability in $\clubsuit$  although it is almost perfect nil hand since the first $\heartsuit$ or $\spadesuit$ trick will remove this vulnerability.  On the other hand, the naive classifier will classify as nil a hand like $5\diamondsuit, 8\diamondsuit, T\diamondsuit, A\diamondsuit, 5\clubsuit, 8\clubsuit, T\clubsuit, 5\heartsuit, 8\heartsuit, T\heartsuit, 5\spadesuit, 8\spadesuit, T\spadesuit$, since each suit is just slightly safer than the vulnerability bar. However, when combining the vulnerability in all four suits, this hand is unsafe.

\subsection{UCT}\label{app:uct}
The UCT algorithm\cite{kocsis2006bandit} is a strong candidate in multiplayer games \cite{sturtevant2008analysis}.  UCT evaluates the best action out of the possible actions by virtually completing the round many times and choosing the action that produce the best result on expectation.  It evaluate the action by it end of the round result, this is important since it is hard to evaluate a state during a round.  
\vfill\eject

\begin{algorithm}[ht]
\SetAlgoLined
\KwResult{card that maximize $E[$point gap$]$}
 initialization\;
 \While{time $<$ TIME LIMIT}{
    Deal unseen cards\;
    Play card that maximize $f(x) = x + C\sqrt{\ln{n}/n_i} $ \;
    \While{trick number $\leq 13$}{
        \While{number of played players $\leq 4$}{
            Play card (heuristic) \;
            Update node data \;
        }
        Update end of trick data \;
    }
    Update $X_{i,t} = points_{partners} - points_{opponents}$
 }
 Return the card with highest $\bar{X}_i$ \;
         
\caption{UCT for the playing phase of Spades} \label{alg:UCT} 
\end{algorithm}
Where, $n$ is the number of iteration UCT made.  $n_i$ is the number of iteration where card $i$ was played. $X_{i,t} $ is the points gap between the partnership and the opponents, and $\bar{X}_i$ is the average $X_{i,t} $.  

\paragraph{UCT Limitations}
Our implementation of UCT suffer from three limitations. First, it needs high computation power however mobile games players prefer short waiting time, thus the app demands the agent to play a card in less than 3 seconds, thus unless the state is the last few tricks of the round, UCT will not make enough iterations in order to find the best action.  Second, it is hard to model the knowledge a player gain about other players' hands from observing their actions.  That is to say, a naive UCT deals the unknown cards uniformly between the three other players, however some actions indicate that the distribution is different.  An obvious example is the case when a player fail to follow suit, that means she has no cards in that suit.  Less obvious example is that high bids means higher probability of holding high cards and spades.  AI-Factory implemented a search algorithm that infer the probability of each card to be in each hand \cite{Opponent2019HandSpades}.  
Lastly, UCT search the best action by playing to the end of the round, however it is not obvious how each player will play.  Our implementation assume the opponents are playing as \SRP would play, without an accurate opponent modeling UCT reach to inaccurate results.  An accurate opponent modeling will improve the results of the UCT.

\paragraph{Comparing \SRP to UCT playing modules}
    UCT is stronger than \SRP when activated in the last few tricks of the rounds.  The reason UCT is better than \SRP only when activated in the last tricks is due to a time restriction per move.  Since the time needed to preform well using \UCT is growing exponentially in the remaining tricks, the amount of time needed to allow \UCT to reach a good bid isn't reasonable, this conclusion was also reached by AI-Factory\cite{whitehouse2013integrating}.

\section{Additional aspects in the bidding}

\subsection{End-of-game Bidding Modifications}\label{subsub:End-of-game}
\paragraph{Risk seeking bids when opponents are winning:}
When the opponents can win the game on this round, \BIS is modifying her bid in four cases: (1) opponents have bid nil, (2) high sum of bids, (3) opponents are winning by small points gap, or (4) opponents are winning by a medium points gap and \BIS has a risky nil hand.

(1) When at least one opponent is bidding nil setting the nil is a possibility.  On those rounds \BIS decreases its bid by $0.5$ in order to have an opportunity to set the nil without jeopardizing her own bid. 
(2) When no player has bid nil and the sum of the four bids (including the agent that now consider to modify her own bid) is at least 11, \BIS modifies her bid such that the sum of bids will be  14.  This guarantee that one of the contracts will be set.  If the opponents fulfill their contract, then they win anyway.  If they do not, then the partners makes enough tricks to fulfill their contract.
(3) When the opponents are going to win the game by a small margin of less than 20 points, \BIS will increase its bid by up to two tricks.  
(4) When the points gap is larger than 20 points and \BIS holds a risky but do-able nil hand, a nil bid will be made.

\paragraph{Risk averse bids when partnership is winning:}
When the points gap is large in \BIS favor, a conservative bid that reduces the chances of being set increases the winning probability.  There are two risk averse bid-modifications: (1) avoid nil bid, and (2) reduce bid by one trick.

(1) When \BIS is in the last position and about to bid a nil bid, however it notice that it can win the game even with it's calculate regular bid it will bid its regular bid.

(2) When \BIS sits in the last position and notice that it will win the game even with a contract smaller by one trick and it can get two bags without receiving the bags penalty, than \BIS will bid one less than its original calculated bid.

\subsection{Signaling Conventions for Nil Bidding}
\BIS uses two bidding conventions when playing with another \BIS partner.  The signal is passing from the first-to-bid partner over the second-to-bid partner.  This means that the first-to-bid partner will only bid the reserved bid if certain conditions are meet.  \BIS uses one of the following two conventions, depending on the current score.  In most rounds the Big 5 convention is on, however when the partnership is behind by more than 100 points, the big 5 is replaced by the big 6 convention.  The reserved bid is not used if the convention criteria is not met, instead \BIS will decrease bid by one.  Using this conventions denies \BIS from bidding a certain bid which can result in a loss of either 9 or 20 points (depends if the bags limit will be reached), however the benefit from succeeding nils at higher rates is larger than the that loss.

\begin{itemize}
\item \textbf{Big 5 - offering nil cover}\\
When first partner is bidding, the `5' bid is reserved to a ``please bid nil" signal.  This bid means that agent holds a good covering hand, which reduces the nil threshold for partner.  The condition for a good covering hand are: (1) $A\spadesuit$ or $K\spadesuit$, (2) four to six takes, (3) in each of the side suits: at least one card higher than ten or void suit.  Note that the $A\spadesuit$ or $K\spadesuit$ condition allows partner to bid nil even if she holds the $K\spadesuit$ or $Q\spadesuit$ which normally deny a nil bid.

\item \textbf{Big 6 - demanding Blind nil}\\
When first partner is bidding and the partnership is behind by a lot, the `6' bid is reserved to signal that she holds a good covering hand, which encourage her partner to bid \textbf{blind-nil}.  The condition for a good covering hand are: (1) $A\spadesuit$ (2) at least four takes, (3) in each of the side suits: Ten or higher or a void.  Note that if the first partner does not hold the $A\spadesuit$, then the second partner holds this sure-nil-lose card with probability $1/3$.
\end{itemize}

\subsection{Blind Nil}
\begin{figure}
\centerline{\includegraphics[width = 300pt, height = 150pt, keepaspectratio] {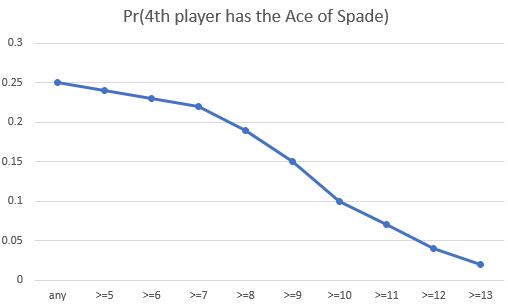}}
    \caption{the probability of the 4th player to hold the $A\spadesuit$ as a function of the sum of bids by the first 3 players.  Gathered from a dataset of several millions games.} \label{fig:4th_player_As}	
\end{figure}

Some variants of the game allow \textit{blind nil} bids which means bidding \textit{nil} before looking at your cards for double the reward/penalty.  Biding blind nil with no information on other players' hands (when bidding first) is too dangerous and results in negative expected points.  Simple observation is that $1/4$ of the hands are doomed to fail due to holding of the $A\spadesuit$.  By examine completed games we found that the probability of holding the $A\spadesuit$ monotonically decrease in the sum of bids of the three previous players (see Figure~\ref{fig:4th_player_As}).  Thus higher sum of bids make it safer to bid blind nil. This method (increasing the probability of bidding blind nil when sum of bids is larger than X) can be used as an alternative to the Big 6 signal convention.  Currently \BIS only bid blind nil if partner signal the Big 6 or in desperate situations\footnote{desperate situations defined as rounds where the opponents will win the game if they make their bid and their bid is easy (no nil's and the sum of bids is below 11)}, however it looks like bidding blind nil in forth position when the sum of bids is high could be beneficial.  This bid should only be played small portion of the times, otherwise opponents will have a beneficial deviation to unrealistic high bid that will persuade \BIS into foolish blind nil bids.

Note that the chance of a random hand to be a `sure lose nil hand' is higher than $25\%$ since there are other holdings that also insure winning a trick, the most obvious ones are in the $\spadesuit$ suit: 
$A\spadesuit$, 
$\{K\spadesuit,Q\spadesuit\}$, 
three $\spadesuit$'s higher than 9, 
four $\spadesuit$'s higher than 7, 
five $\spadesuit$'s higher than 5,
six $\spadesuit$'s higher than 3, or any
seven $\spadesuit$'s.

\section{Terminology}\label{app: Terminology}
\begin{itemize}
\item bag	(n.) - see overtrick, (v.) To force the opposing team to take one or more overtricks.
\item bag back	(v.) - To accumulate 10 overtricks.
\item bid	(n.) - A declaration to take a certain number of tricks.
\item blind nil	(n.) - A nil bid made before looking at one's hand.  Worth double the points of a normal nil.
\item{boss} - The highest unplayed rank of a particular suit. For example, if the A$\heartsuit$ has been played on a previous trick, the K$\heartsuit$ is the boss.
\item{break spades}	(v.) - To play a spade for the first time in a particular round.  Allows leading spades in the following tricks.
\item{coverer} - niler's partner.
\item{cut}  (n.) - trump, ruff.  Play a spade card in a trick when the leading card is not a spade.
\item{doubleton} - a suit with exactly two cards.
\item{duck}	(v.) - see play under.
\item{dump}	(v.) - To play a card that avoids taking a trick. Usually implies the played card could have taken a future trick. For example, if the A$\clubsuit$ is played, playing the K$\clubsuit$ (instead of a lower club) or the A$\diamondsuit$ is considered dumping.
\item{follow} (v.) - to play after some one already played in that trick.  As apposed to \textit{leading} a trick.
\item{follow suit}	(v.) - To play a card of the same suit as the led suit.  Players must follow the leading suit if they can.
\item{lead}	(v.) - to play the first card of a trick.  (n.) 1. the first card of a trick.  2. the state of being the player who is to play the first card of a trick.
\item{nil} - a bid of 0.  
\item{niler} - a player which bid nil. 
\item{overtrump}	(n.) - To play trump of a higher rank than one's opponent's trump. 
\item{play over}	(v.) - To play a higher ranking card than the current high card.
\item{play under}	(v.) - To play a card of a rank lower than the highest currently played card, or a card of a non-trump suit other than the lead suit.
\item{singleton} - a suit with exactly one card.
\item{spade tight} - a hand containing nothing but spade cards.
\item{trump} - a card of the strongest suit. In Spades, spades are always trumps.
\item{void} - a suit with no cards.
\end{itemize}

\end{document}